\documentclass[
	a4paper, 
	10pt, 
	unnumberedsections, 
	twoside, 
]{LTJournalArticle}

\usepackage{abstract} 

\runninghead{} 

\footertext{Control and Monitoring of Artificial Intelligence Algorithms} 

\title{Control and Monitoring \\ of Artificial Intelligence Algorithms} 

\author{%
	Belén Muñiz Villanueva\thanks{E-mail: bmuniz@mondragon.edu}, Carlos Mario Braga Ortuño\thanks{E-mail: cbragaor@gmail.com} \cr and Blanca Martínez Donoso\thanks{E-mail: blanca.mdonosox@gmail.com}
}

\begin{document}

\maketitle 
\begin{abstract}
This paper elucidates the importance of governing an artificial intelligence model post-deployment and overseeing potential fluctuations in the distribution of present data in contrast to the training data. The concepts of data drift and concept drift are explicated, along with their respective foundational distributions. Furthermore, a range of metrics is introduced, which can be utilized to scrutinize the model's performance concerning potential temporal variations.
\end{abstract}

\section{Introduction}

Upon deploying an Artificial Intelligence algorithm based on Machine Learning for generating predictions, it is crucial to address the potential risk associated with changing environmental or social factors. Despite thorough training, rigorous bias-checking, and successful performance in the test phase, the model may encounter issues when the data it was trained on no longer reflects the current reality accurately. 

This discrepancy between the Training Data Distribution (TDD) and the Live Data Distribution (LDD) can lead to performance degradation, biased predictions, and a deviation from the model's initial explainable criteria. To mitigate these problems, it is essential to compare the distributions, measure their dissimilarity, monitor relevant metrics, and implement a system that uses Control Limits to identify when significant changes in environmental conditions have occurred. This system serves as a valuable tool for determining when the model's life cycle needs to be revisited and retraining should be considered.

A model is considered in which a set of independent variables, denoted as $X_1, X_2, \ldots, X_n$, aims to predict a dependent variable, denoted as $Y$. The $n$-dimensional variable, $X$, is formed by the aforementioned independent variables, and $f_X$ represents the probability density function of this $n$-dimensional variable. On the other hand, $f_Y$ represents the probability density function associated with the variable $Y$ that is being predicted.

In real-world scenarios, the objective is to predict the behavior of the variable $Y$ at a specific moment in time based on known behaviors of the $n$-dimensional variable $X$ at the same moment. Consequently, in such problems, the behavior exhibited by the joint probability distribution of the variables $X$ and $Y$ becomes crucial.

The joint probability density function of the variables $X$ and $Y$, denoted as $f_{X,Y}$, considers $X$ as an $n$-dimensional variable. 

This multidimensional variable takes on specific values defined by the equations.
\begin{itemize}
    \item For discrete statistical variables:
\begin{equation}
\begin{split}
f_{X,Y}(x,y) &= P(X=x, Y=y) \\
&= P(Y=y|X=x) P(X=x) \\
&= P(X=x|Y=y) P(Y=y)
\end{split}
\end{equation}
        Where $\sum_{x=1}^{X} \sum_{y=1}^{Y} P(X=x, Y=y) = 1$.

        \item For continuous statistical variables:
\begin{equation}
\begin{split}
f_{X,Y}(x,y) = f_{X|Y}(x|y) \cdot f_{Y}(y) \\
= f_{Y|X}(y|x) \cdot f_{X}(x)
\end{split}
\end{equation}
        Where  $\int \int f_{X,Y}(x,y) \, dx \, dy = 1$.
\end{itemize}

The simultaneous involvement of the $n+1$ random variables, referred to as $X$ and $Y$, in the prediction event underscores the significance of understanding the joint probability. This understanding becomes crucial when assessing potential distributional changes that may have occurred between the training data distribution (TDD) and the point in time at which a new prediction is desired (LDD). Although our focus will be on the discrete case, it is worth noting that the logical approach remains the same for the continuous case.

The endeavor to ensure that a model predicts from data that exhibits a distribution akin to the training data relies on two fundamental conditions, as illustrated by the following expression:

\begin{equation}
\begin{split}
f_{X,Y}(x,y) &= P(X=x, Y=y) \\
&= P(Y=y|X=x) P(X=x)  
\end{split}
\end{equation}

Or in simplified form:
\begin{equation}
P(X, Y) = P(Y|X) P(X)
\end{equation}

This is, on the one hand, the distribution of the predictor or independent variables, also called the input distribution, and on the other hand, the distribution of the dependent variable or to be predicted conditioned on the predictors or independent ones, also called the distribution of the output of the model conditioned to the input. The satisfaction of these conditions helps ascertain that the model is founded on data reflecting the current reality, unaffected by economic, social, or other contextual changes.

As a result, when comparing the joint distribution during the training phase, referred to as TDD, with the joint distribution during the operational phase, denoted as LDD, any alteration observed in these distributions can be attributed to two factors: either $P_{TDD}(X) \neq P_{LDD}(X)$ or $P_{TDD}(Y|X) \neq P_{LDD}(Y|X)$.

Consequently, these factors represent discrepancies between the input distribution and the conditional output distribution, corresponding to the respective differences between the two temporal moments. By decomposing these factors, the concepts of Data Drift and Concept Drift can be introduced.

The concept of Data Drift encompasses scenarios where the input distribution undergoes changes. It refers to the occurrence of the first factor identified earlier, namely, $P_{TDD}(X) \neq P_{LDD}(X)$. In such instances, there are observable differences between the marginal probability distribution of the independent variable during the training phase (TDD) and the subsequent production or monitoring phase (LDD).

An illustration of this scenario could arise in a hypothetical scenario where one of the variables in the $n$-dimensional variable $X$, denoted as $X_j$, represents a categorical variable indicating the store category in which individuals make their primary monthly purchase. During an economic upheaval, such as a crisis, this categorization can undergo changes, resulting in previously less common options during training becoming increasingly prevalent over time. Consequently, the inputs during algorithm monitoring in the variable $X_j$ predominantly encompass instances that were unrepresentative during the training phase. This situation entails a potential risk where the algorithm may commence generating predictions that lack coherence, as it has not been trained on data that adequately captures the nuances of the current economic landscape. 

The term Concept Drift is employed to describe the second scenario where a change occurs in the distribution between the training and monitoring phases: $P_{TDD}(Y|X) \neq P_{LDD}(Y|X)$. In this case, the conditional distribution of the dependent variable given the independent variable exhibits disparities between the TDD and LDD periods.

An illustrative example of this scenario can be observed in a model that predicts attendance at cultural events based on an input variable indicating the geographic area. In a hypothetical scenario characterized by substantial growth and development in that specific geographic area, shifts in population distribution may occur. For instance, there could be an increase in the influx of young individuals relocating to that area for employment or residential purposes. Although the input variable, denoted as $X_j$, remains unchanged (representing the geographic area), and the target variable (attendance at cultural events) remains constant, the proliferation of individuals from this new population, which has contributed to the significant development of the geographic area, can lead to shifts in preferences and cultural tastes over time.

This demonstrates how, in the case of Data Drift, changes are identified in certain input variables of the model, rendering them no longer representative of the current reality. On the other hand, in the case of Concept Drift, the input variables remain unaltered. However, alterations in monitoring conditions compared to those during the training phase indicate that the model no longer accurately reflects the current reality.

Consequently, the objective of this document is to present techniques for measuring the disparities between the training data distribution (TDD) and the monitoring data distribution (LDD).

To achieve this objective, the document is structured into the following points:
\begin{itemize}
    \item The introduction of the European regulatory framework, which, along with ethical aspects, justifies the need for monitoring.
    \item The introduction of fundamental concepts that provide support throughout the document.
    \item Techniques for measuring Data Drift will be explored.
    \item Techniques for detecting Concept Drift will be examined.
    \item A monitoring model will be proposed to identify the necessity of retraining the model to adapt its behavior to the new environment.
    \item A summary of conclusions will be presented, which aims to justify and clarify the need for monitoring models based on Machine Learning.
\end{itemize}

\section{Regulatory and Normative Aspects}

Regarding regulations, there is current legislation that deals with responsibility and accountability in the event of causing damage, as well as recommendations from both legislatures and regulatory bodies in some industries.

Among the main regulations applicable in the countries of the European Union, several key legislations can be identified. It is important to highlight the following:
\begin{itemize}
    \item Article 38 of the Charter of Fundamental Rights of the European Union \cite{EUCharter}; guaranteeing high consumer protection in Union policies.
    \item Articles 4.2.f, 12, 114 and 169 of the Treaty on the Functioning of the European Union (TFEU) \cite{EUTreaty}, which establish the responsibility in terms of consumer protection.
    \item Council Directive (EU) 85/374/EEC of July 25, 1985 \cite{EUDirective}, regarding liability for damages caused by defective products
\end{itemize}

Recently, on 09/28/2022, the European Commission, under the name of \textit{"New liability rules on products and AI to protect consumers and foster innovation"} \cite{EULiabilityRules}, has published a proposal to modernize the existing rules on the liability of manufacturers on defective products, including products based on AI, in order to establish a legal security framework that invites companies to innovate and customers to consume these products being aware of their rights.
Additionally, and in the form of a proposal, since 2021 there is the AI law (IA Act) that will see the light in 2024 and that will formalize a regulatory framework with the aim of favoring a sustainable and beneficial development of AI for people, which will reinforce the use responsible for products based on Artificial Intelligence.
In terms of recommendations, the following points can be highlighted:
\begin{itemize}
    \item Ethics of Artificial Intelligence Recommendation \cite{UNESCOAI}, approved by UNESCO on November 23, 2021 where member states are urged to ensure that AI models benefit people and therefore ensure that results of the models do not harm or discriminate against people; something highly likely to happen when the productive model is not dealing with a similar environment to the one in which it was trained.
    \item European Parliament resolution of 20 October 2020, with recommendations to the Commission on a civil liability regime for Artificial Intelligence (2020/2014(INL) \cite{EUCivilLiability}.
    \item As part of the Basel recommendations for risk models, propose the adoption of various techniques and metrics. These recommended approaches aim to enhance the assessment and monitoring of risk models. Two specific techniques that merit attention are the Brier Score and the Kolmogorov-Smirnov Test/Divergence \cite{BancoEspEstabilidad}.
\end{itemize}

\section{Main Concepts}

\subsection{Kullback-Leibler Divergence}
Along this paper Kullback-Leibler divergence \cite{ChaaHistograms}, will be used as measure to assess the similarity between two probability distribution functions. This metric, introduced by Salomon Kullback and Richard Leibler in 1951, is known as Kullback-Leibler divergence, information divergence, information gain or relative entropy.
Given two probability distributions P and Q for two random variables, the K-L divergence is defined as the weighted average of the logarithmic difference between probabilities P and Q.
In the discrete case:
\begin{equation}
D_{K-L}(P|Q) = \sum P(i) \ln\left(\frac{P(i)}{Q(i)}\right)   
\end{equation}
Or in the continuous case:
\begin{equation}
D_{K-L}(P|Q) = \int P(x) \ln\left(\frac{P(x)}{Q(x)}\right) \, dx  
\end{equation}

In other words, the Kullback-Leibler divergence is a measure of information loss or difference between two probability distributions. It quantifies the additional information needed to transform one distribution into another, thus providing a measure of discrepancy or dissimilarity between them.
The main properties of this divergence are:
•	It takes the value 0 only when P and Q are equal.
•	In the remaining cases it always takes positive values.
•	This measure is not considered as a distance because it is not symmetrical, that is; It is not always true that $D_{K-L}(P|Q) = D_{K-L}(Q|P)$.
In statistics this measure is closely linked to the method of estimating the parameters of a presumed probability distribution through the maximum likelihood method \cite{ShiensKL}. It will also be used later to define distances (symmetric measurements) between distributions. Both metrics will enable the assessment of the similarity or dissimilarity between the TDD distribution and the LDD distribution.

\subsection{Data Binning}
Data Binning, also known as Data Discretization, is a data processing technique employed to partition a range of data into distinct containers or bins. Within each interval, a representative value is chosen to summarize the data (typically a measure of central tendency like the mean or median). This technique is commonly utilized to mitigate the impact of observation errors.

Data Binning enables the treatment of a continuous distribution sample as a discrete distribution that encompasses as many values as the selected number of bins or intervals for discretization. From these groups, empirical distribution functions of the random variables can be constructed based on either the training or evaluation data.

\subsection{Kolmogorov-Smirnov Test}
The Kolmogorov-Smirnov statistic or KS distance \cite{CristobalInferencia}, is a measure to evaluate the similarity between two data distributions. It’s calculated as the maximum absolute distance between the cumulative distribution functions of two samples. It’s mostly used to determinate how similar is a data set from a theorical distribution. The range of values that the KS distance can take is between 0 and 1. The smaller the value, the closer the compared distributions are. This statistic has the characteristic of being sensitive to differences in both the form and location of the accumulated empirical distribution functions.
Although the KS statistic is not a metric for monitoring Drift, its use can be interesting in classification problems. Thus, let’s consider a set of explanatory variables ($X1, X2, \ldots, Xn$) and generate distributions conditioned on the different classes of the predicted variable. 
In this situation, the KS statistic is useful for comparing the conditioned distributions and determining their level of similarity for the following reasons:
\begin{itemize}
    \item On the one hand, it helps to determine if a variable is contributing differently between the two groups.  This behaviour is a good leading indicator of a potential degradation in the predictive performance of the model since it is reasonable to expect better prediction performance for greater distance between groups.
    \item On the other hand, KS can be used to compare each variable from the Training Data Distribution (TDD) and from the Live Data Distribution (LDD). In other words, KS can be used to analyse when the distance between two groups in the TDD data is similar to the distance between two groups of the LDD data. In case of changes in behaviour or tendency over the time, KS will be a leading indicator to consider that something has vary regarding the training data. This performance indicates the need to review other metrics to evaluate the need of retraining.
\end{itemize}


\section{Data Drift}
As mentioned above, the validity of a model heavily depends on the similarity between the distribution of the data on which it was trained (TDD) and the distribution of the data used to predict at any given time (LDD). When these distributions differ significantly, it should be assumed that the current reality is not the reality that the data used to train the model showed, and it is necessary to responsibly consider retraining the model. This will avoid the algorithm inferring results that could negatively affect the existing performance and neutrality metrics at the time it was put into production.
Therefore, this study presents different complementary approaches that will allow us to measure and monitor if a model is affected by Data Drift.

\subsection{Covariate Drift}
Adhering to the notation criteria followed thus far, the training data distribution will be identified as TDD, while the distribution of the data intended for prediction at a given time point will be denoted as LDD. The primary objective is to determine the degree of similarity or difference between these two distributions.

The Covariate Drift metric is defined as the non-intersection distance between the TDD and LDD distributions and it is formally expressed as:
\begin{equation}
d(\text{TDD, LDD}) = 1 - \sum \min(\text{TDD}_i, \text{LDD}_i)
\end{equation}
To perform this calculation, for each independent variable, the data is grouped into equally sized bins, starting from the range of values of the independent variable and considering both TDD and the LDD data. Usually, around 20 intervals per independent variable are considered.  

Afterwards, the relative frequency is calculated for all the intervals of the Training Data (TDD) and the Live Data (LDD), considering both distributions separately. Then, for each interval, the minimum of these relative frequencies is selected. For each given independent variable, the sum of the selected minimum frequency for each interval must be subtracted from 1 to obtain the distance defined as Covariate Drift. 

The \textbf{thresholds} usually identified on the Covariate Drift metric to evaluate the presence of Data Drift are: 
\begin{equation}
\begin{aligned}
&\text{If } d(\text{TDD, LDD}) > 0.4 \rightarrow \text{High Drift} \\
&\text{If } 0.3 < d(\text{TDD, LDD}) \leq 0.4 \rightarrow \text{Medium Drift} \\
&\text{If } 0.2 < d(\text{TDD, LDD}) \leq 0.3 \rightarrow \text{Low Drift} \\
&\text{If } d(\text{TDD, LDD}) \leq 0.2 \rightarrow \text{Non-existing Data Drift}
\end{aligned}
\end{equation}
The simplicity of its calculation is a positive point for its adoption.
However, the outcome is heavily contingent upon the quantity and magnitude of the intervals, thereby necessitating the use of complementary methods for drift measurement. Subsequently, in the subsequent sections, alternative techniques for assessing drift will be explored, offering a more comprehensive evaluation of the evolving data patterns.

\subsection{Wasserstain Distance}

The Wasserstein distance is based on a concept introduced in 1781 by Gaspard Monge \cite{WikipediaWasserstein}, a French mathematician who studied the optimal transport of resources. Due to the context in which the metric was originated, this distance is also known as Earth Mover's Distance (EMD).
The original problem assumed the existence of two locations, $P$ and $Q$, in a construction project, where there were m piles (an element used for the foundation of works) at point $P$ and $n$ pits at point $Q$. Under this assumption, the goal was to study how to move the $m$ piles from point $P$ to the $n$ pits at point $Q$ while minimizing the work done.

Considering the possibility of multiple piles being placed within a single construction pit, the position and width of each pile are denoted by the pair ($p$,$w$), while the position and width of each pit are denoted by the pair ($q$,$w$). Thus, the sets $P$ and $Q$ can be represented as follows:
\begin{equation}
P = {(p_1, w_{p_1}), (p_2, w_{p_2}), \ldots, (p_m, w_{p_m})}
\end{equation}
\begin{equation}
Q = {(q_1, w_{q_1}), (q_2, w_{q_2}), \ldots, (q_n, w_{q_n})}
\end{equation}

By denoting the distance between pile $p_{i}$ and pit $q_{j}$ as $d_{ij}$, a distance matrix $D$ of dimensions $m$ x $n$ can be constructed:
\begin{equation}
D = [d_{ij}]
\end{equation}
The objective is to determine a flow that can be represented by the matrix $F$:
\begin{equation}
F = [f_{ij}]
\end{equation}
where $f_{ij}$ denotes the flow between pile $p_{i}$ and pit $q_{j}$. The matrix $F$ represents the allocation of piles to pits based on the distances calculated in the matrix $D$. $f_{ij}$ takes the value 1 if the pile i is moved to pit j or the value 0 otherwise, in order to minimize the function representing the workload:
\begin{equation}
W(P,Q,F) = \sum \sum d_{ij} f_{ij}
\end{equation}

The \text{EMD} measure is the normalization of the workload: 
\begin{equation}
\text{EMD}(P,Q,F) = \frac{\sum \sum d_{ij} f_{ij}}{\sum f_{ij}}
\end{equation}
The Wasserstein distance is the translation of this idea to a scenario in which there are two probability distributions: one for the independent variables of the test data and, another one, for the independent variables of the data being used to make the prediction. The purpose of the metric, is to calculate the minimum amount of work required to transform one distribution into the other. It has the particularity of not requiring data grouping and the algorithm for its calculation, simplifying the case of two one-dimensional distributions, would be as follows: 
\begin{enumerate}
    \item Given the following input data:
$TDD = {x_0,\ldots, x_{\mathrm{n-1}}}$
$LDD = {y_0,\ldots,y_{\mathrm{n-1}}}$
\item Initialize $w_0 = 0$ 
\item Iterate from $i=1$ to $n$ and calculate: $w_i = x_{i-1} - y_{i-1} + w_{i-1}$
\item The Wasserstein distance is obtained by summing the absolute values of $w_i$:
\begin{equation}
\text{EMD}(\text{TDD, LDD}) = \sum |w_i|
\end{equation}

\end{enumerate}
Although it is not the focus of this document, it is worth mentioning that the above algorithm is equivalent, and therefore gives the same result, as computing the Manhattan distances on the cumulative distributions.

\subsection{Stability Index (CSI, PSI)}

The Stability Index metric quantifies the population change in a variable, which can be either an independent/predictor variable or a dependent/target variable. It measures the percentage change in population over the time interval between training (TDD) and the specific prediction moment (LDD).

Therefore, this metric can be decomposed into two metrics:
\begin{itemize}
    \item PSI (Population Stability Index), which measures the percentage change over time for a predictor variable.
    \item CSI (Characteristic Stability Index), which measures the change over time for the predicted variable.
\end{itemize}

The premise on which these metrics are based is that a prediction model works better if the data used to train does not differ too much from the data on which it is validated. These metrics are limited to a single variable, as the dependent variable to be predicted is dependent on the predictor variables, , so the CSI metric can be interpreted as a measure of how the entire dataset that the algorithm is working with is changing.
Therefore, both PSI and CSI can help us to identify changes that may affect the model and worsen its performance. To calculate these metrics, it is necessary to resort to data grouping techniques. The focus will be on the discrete case for their development, utilizing the Kullback-Leibler divergence.
Considering TDD as the probability distribution function of the data at the time of training and LDD as the probability distribution function of the data at the  prediction time, it can be formally defined the Stability Index (SI) metric as follows:
\begin{equation}
\begin{split}
SI(TDD, LDD) = D_{K-L}(TDD|LDD) + \\
D_{K-L}(LDD|TDD)
\end{split}
\end{equation}
Given that:
\begin{equation}
D_{K-L}(P|Q) = \sum P(i) \ln\left(\frac{P(i)}{Q(i)}\right)
\end{equation}
the Stability Index (SI) metric can be expressed as follows:
\begin{equation}
\begin{aligned}
SI(TDD, LDD) = \sum TDD(i) \ln\left(\frac{TDD(i)}{LDD(i)}\right) + \\ 
\sum LDD(i) \ln\left(\frac{LDD(i)}{TDD(i)}\right)
\end{aligned}
\end{equation}
or equivalently:
\begin{equation}
\begin{aligned}
SI(TDD, LDD) = \sum TDD(i) \ln\left(\frac{TDD(i)}{LDD(i)}\right) - \\ 
\sum LDD(i) \ln\left(\frac{TDD(i)}{LDD(i)}\right)
\end{aligned}
\end{equation}
Taking out a common factor, the SI metric is written by:
\begin{equation}
\begin{aligned}
SI(TDD, LDD) = \\
\sum (TDD(i) - LDD(i)) \ln\left(\frac{TDD(i)}{LDD(i)}\right)
\end{aligned}
\end{equation}

From an algorithmic point of view, it would be calculated as follows:
\begin{enumerate}
    \item Together considering the data at TDD and LDD moments, group the data and construct intervals of equal length for non-categorical variables. Ten intervals are sufficient, and categorical variables can be left ungrouped, considering each value of their range of possible values as an interval.
    \item For each interval, calculate the relative frequency of values in that interval, separately for both TDD and LDD data.
    \item For each interval, calculate the difference of observed percentages between TDD and LDD.
    \item For each interval, the ratio of observed percentages between TDD and LDD data is calculated.
    \item For each interval, the result obtained in step 3 is multiplied by the natural logarithm of the result obtained in step 4.
    \item Finally, sum all the results obtained in step 5 to obtain the Stability Index metric.
\end{enumerate}

It is also noteworthy that CI and therefore PSI and CSI are symmetric metrics and can be treated as distances; that is: $SI (P,Q) = SI (Q,P)$
Finally, it should be mentioned that if the evaluation sample has exactly the same distribution of values as the construction sample, stability will be maximum and the value of the index will be 0. Below is the empirical norm for determining to what extent a variable has changed distribution:
\begin{equation}
\begin{aligned}
&\text{If } SI < 0.1 \rightarrow \text{Very slight change} \\
&\text{If } 0.1 < SI \leq 0.2 \rightarrow \text{Not significant change} \\
&\text{If } SI \leq 0.2 \rightarrow \text{Significant change}
\end{aligned}
\end{equation}

\subsection{Jensen-Shannon Distance}

Another way to measure the similarity or difference between TDD and LDD distributions is the Jensen-Shannon distance. The Jensen–Shannon divergence \cite{EndresJensenShannon}, is calculated from the Kullback–Leibler divergence as follows:
\begin{equation}
\begin{aligned}
D_{J-S}(TDD|LDD) = \\
\frac{D_{K-L}(TDD|M) + D_{K-L}(LDD|M)}{2}
\end{aligned}
\end{equation}
Where the random variable $M$ is calculated as:
\begin{equation}
M = \frac{TDD + LDD}{2}
\end{equation}
The Jensen-Shannon divergence can take values between 0 (there is no change between the distributions) and 1 (highly significant change in the distributions).
The Jensen-Shannon distance (JSD) is the square root of the divergence, that is:
\begin{equation}
\text{JSD}(TDD | LDD) = \sqrt{D_{J-S}(TDD | LDD)}
\end{equation}

This distance is based on the concept of Mutual Information, which belongs to the field of Information Theory. Mutual Information quantifies the information obtained from one random variable when another random variable is known. The Jensen-Shannon distance is derived from this concept by considering a weighted combination of the mutual information metrics between the random variables TDD and LDD with respect to a third random variable, which is calculated based on a mixture or weighting of both variables. The symmetrical nature of the Jensen-Shannon divergence allows us to define a distance measure using this divergence.

However, from a practical perspective, this approach is less useful compared to the Stability Index metric. This is because it involves the calculation of a new distribution, denoted as M, based on the TDD and LDD distributions.


\section{Concept Drift}

In addition to the Data Drift, it may happen that without changing the data distribution of the independent or predictor variables, the interpretation of these data respect to the dependent variable or to predict changes, and this phenomenon is known as Concept Drift.
$PTDD (Y|X)\neq PLDD (Y|X)$

The measurement of Concept Drift is more complex compared to Data Drift. It necessitates knowledge of both the predicted value $\hat{Y}$ and the true value of the prediction $Y$. However, the actual value $Y$ is typically unknown to the models since the objective is to predict it.
Therefore, the measurement of the Concept Drift requires an effort to collect current data – from the moment of prediction -, in which the real value of the dependent variable (the one to be predicted) is known.

\subsection{Page-Hinkley Test}

The Page-Hinkley test or PHT is a widely used metric to detect the Concept Drift, since it can be used for both classification and regression problems \cite{MoussPageHinkley}.
It is based on the error rate of the model $(y - \hat{y})$ and, in particular, on the calculation of the average of these rates. This way, the metric combines the drift of the current moment and the historical vision through a measure of central tendency of the past drift.
Let’s denote $e_t$ as the error at a given time $t$, and $\hat{e}_t$ the average of all the errors up to that time $t$; I mean:
\begin{equation}
\hat{e}_t = \frac{1}{t} \sum_{i=1}^{t} e_i
\end{equation}
Let $\alpha$ be the tolerance to change, so the accumulated error rate can be calculated as:
\begin{equation}
m_t = \sum_{i=1}^{t} (e_i - \hat{e}_t - \alpha)
\end{equation}
As the calculations of $m_1, m_2,\ldots, m_{t-1}$ and $m_t$ are performed over time, it is also possible to determine their minimum value using the following approach:
\begin{equation}
M_t = \min_{j=1,\ldots,t} (m_j)
\end{equation}
The idea behind the Page-Hinkley test (PHt) statistic; is to compare at each time t the values of mt and Mt and show an alert when the difference between mt and Mt exceeds a certain threshold, because that represents an increase in the mean of the error. By applying a similar approach but calculating the maximum instead of the minimum, it is possible to detect decreases in the mean error. In both cases, considering the error at the training time, a threshold $\lambda$ can be established that triggers an alarm when $m_t - M_t \geq \lambda$.

However, in order to calculate the final value of the PHt statistic, an adjustment is made to assign more weight or importance to the most recent measurements compared to the older ones. This adjustment involves introducing the $L_t$ value, which is calculated as follows:
$L_0 = 0$
\begin{equation}
\begin{aligned}
L_j = \left(\frac{j-1}{j}\right) L_{j-1} + \left(e_j - \hat{e}_j - \alpha\right) \quad \text{for } j = 1,\ldots,t
\end{aligned}
\end{equation}
Then, calculate the minimum over these weighted errors:
\begin{equation}
M_t = \min_{j=1,\ldots,t} (L_j)
\end{equation}
To finally calculate the Page-Hinkley statistic at a time $t$ as:
\begin{equation}
P_{G_t} = L_t - M_t
\end{equation}
Since $(j-1)/j$ increases as $j$ increases, greater weight is assigned to the most recent error rates.
Once the alarm threshold $\lambda$ has been set (could be set from the known error in the training phase), in cases where $PG_t \geq \lambda$ it would be convenient to review and verify if the model needs retraining. It is important to consider initializing the measurement process after each retraining and not to consider the previous errors.

\subsection{Brier Score}

This metric is, for its simplicity of calculation and ease of interpretation, one of the most used metrics for the assessment of the goodness or performance of models. In fact, it appears as a recommendation to value the credit risk models in the standards created from the Basel agreements.
It can be used for any model that assigns an output based on a probability, it is given by the mean square error between the predicted and actual probability. 
\begin{equation}
BS = \frac{1}{N} \sum_{i=1}^{N} (y_i - \hat{y}_i)^2
\end{equation}
where $Y$ is the actual probability, $\hat{Y}$ is the probability predicted by the algorithm and $N$ is the number of cases that have been computed for the prediction.
This metric always takes values between 0 and 1, being 0 the value that indicates a good accuracy in the prediction , while higher values indicate a greater decay in the performance of the model.

\subsection{Early Drift Detection Method (EDDM)}

For a gradual degradation in the model, it is complicated to detect and determine when there is Concept Drift problem based uniquely in the metrics mentioned above. The EDDM or Early Drift Detection Method technique tries to alleviate this problem \cite{BaenaGarciaDrift}.  Since it is based not on the measurement of the magnitude of the errors $(Y-\hat{Y})$, but on the measuring of the distance between the errors occurred in two consecutive measurements. 

EDDM operates by monitoring the errors or residuals produced by the model when making predictions on incoming data. It calculates the mean and standard deviation of these errors over a sliding window of recent instances. The mean and standard deviation are updated incrementally as new instances arrive, allowing EDDM to adapt to changes in the data. 

The key idea behind EDDM is to compare the current error statistics with the historical values. If the current error exceeds a certain threshold, it suggests a potential concept drift. EDDM uses statistical hypothesis testing to determine if the observed change in error statistics is significant enough to declare a drift. 

From the previous definition, the $EDDM$ Score is calculated as follows: 
\begin{equation}
EDDM_i = \frac{\mu_i + 2\sigma_i}{\mu_{\text{max}} + 2\sigma_{\text{max}}}
\end{equation}
Where $\mu_i$ and $\sigma_i$ are respectively the mean and standard deviation between error distances and $\mu_\text{max}$ and $\sigma_\text{max}$  represent each the maximum values of $\mu_i$ and $\sigma_i$. 

The following criterion is usually used to determine when there is a Model Drift problem: 

\begin{equation}
\begin{aligned}
&\text{If } EDDM_i > 0.95 \rightarrow \text{No alarm should be triggered} \\
&\text{If } 0.90 < EDDM_i \leq 0.95 \rightarrow \text{Reviewing must be} \\
&\text{carried out} \\
&\text{If } EDDM_i \leq 0.90 \rightarrow \text{Action to avoid drift and } \\
&\text{reset the maximum}
\end{aligned}
\end{equation}

Intuitively, the $EDDM$ method calculates the ratio between the average error and the maximum error. This ratio aims to detect cases where the oscillations leading to a new maximum are increasing in relation to the average error. If the error rate has remained stable over time, the emergence of new maximums in that error will cause a decrease in the value of the $EDDM$ metric. If that decrease crosses the previously established thresholds, it will be time to evaluate whether this situation is justified by a genuine drift problem or by an increase in noise in the data (false positive).  

\subsection{Hierarchical Linear n Rate (HLnR)}

The previous discussions have primarily focused on metrics that address the impact of distribution changes on general error rates or algorithm accuracy. However, there are scenarios where this approach may not be suitable. For instance, problems involving unbalanced data may prioritize metrics other than accuracy, such as reducing false positive rates. Additionally, monitoring algorithmic fairness metrics like additive counterfactual fairness (ACF) or counterfactual unfairness (CUF) can be of interest in certain cases.

To address these situations, a new method for measuring Concept Drift is introduced, which relies on monitoring key metrics derived from the Confusion Matrix, encompassing both performance and neutrality aspects. This method is applicable only in cases where the Confusion Matrix can be generated. Before delving into the calculation of the HLnR (Holding the Line Rate), let us revisit the Confusion Matrix and highlight some of the key metrics that will form the basis of this calculation.

\begin{table}[ht]
    \centering
    \caption{Confusion Matrix}
    \begin{tabular}{|p{1.8cm}|p{2.3cm}|p{2.3cm}|}
        \hline
        \textbf{Actual / Predicted} & \textbf{0(Negative)} & \textbf{1(Positive)} \\
        \hline
        0(Negative) & TN: True Negatives & FP: False Positives \\
        \hline
        1(Positive) & FN: False Negatives & TP: True Positives \\
        \hline
    \end{tabular}
\end{table}

The HLnR (Holding the Line Rate) approach does not involve recalculating the Confusion Matrix and its associated metrics. Instead, it relies on leveraging an existing Confusion Matrix and its n pre-calculated metrics. The objective is to monitor the change in each metric of interest following each new prediction. The calculation of this "change" will be presented as follows \cite{YuConceptDrift}.

Let $R_t$ denote the value of any specific metric (e.g., TPR, TNR, PPV, NPV, etc.) at time $t$.

The proposal is to calculate $R_t$ as follows from the data in $t-1$:
\begin{equation}
R_t = \eta_{t-1} \cdot R_{t-1} + (1 - \eta_{t-1}) \cdot \lambda
\end{equation}
where parameter $\lambda$ takes value 1 if $Y = \hat{Y}$  in $t-1$ and takes value 0 in another case, $\eta$ takes a value in (0,1) and it is called “time decay” and  $R_{t-1}$ as well as $R_t$ complies with the rule:
$R_{t-1} \in (\text{TPR, TNR, PPV, NPV, etc.})$

Therefore $R_t$ is updated using a linear combination of the past value and the "time decay" factor.
It is important to note that the update of $R_t$ occurs only when there is a change in the specific metric being studied. If the focus is on monitoring the True Positive Rate (TPR), for example, the metric will only be updated following a prediction in cases where a True Positive ($Y=1$ and $\hat{Y}=1$) or a False Negative ($Y=1$ and $\hat{Y}=0$) occurs. In all other cases, the value of $R_t$ remains unchanged, i.e., $R_t = R_{t-1}$.

It is necessary that the "time decay" factor provides knowledge about the current performance of the model and the change in terms of concept drift over time, so for its calculation weights are used so that a greater weight is assigned when the current value of the metric is greater than the previous one and lower weight otherwise.
In this way the "time decay" factor is calculated as follows:
\[
\eta_t = \begin{cases}
    (\eta_{t-1} - 1) \cdot e^{-(R_t-R_{t-1})} + 1,  \text{ if } R_t \geq R_{t-1} \\
    (1 - \eta_{t-1}) \cdot e^{(R_t-R_{t-1})} + (2 \cdot \eta_{t-1} - 1), \\
    \text{ if } R_t < R_{t-1}
\end{cases}
\]

By calculating $R_t$ and $\eta_t$, one can proceed to determine $R_{t+1}$ after a new prediction, using the values of $R_t$ and $\eta_t$. Once $R_{t+1}$ is obtained, $\eta_{t+1}$ can be calculated in a similar manner as discussed before. Continuously monitoring the trends of $R$ and $\eta$ over time enables the identification of an alarm threshold. If the values exceed this threshold, it indicates the need for action, such as retraining the model, and resetting the calculations of $R$ and $\eta$ to track the updated performance.


\section{Monitoring Metrics}

In the context of metric monitoring, the utilization of control diagrams, similar to those employed in Statistical Quality Control \cite{HikawaControlCalidad}, can be advantageous. The lower control limit consistently assumes a value of 0, while the upper control limit represents the predetermined threshold for each metric, signifying the maximum acceptable value in our case when dealing with distances and divergences.

Despite their industrial origin, parallels can be drawn between control charts and the relationship between an algorithm and a production machine for predictions. By employing control methods, the temporal progression of metrics can be assessed, enabling the identification of both Data Drift and Concept Drift.

Consequently, this approach facilitates the visualization of metric evolution, the detection of degradation patterns, and the implementation of proactive measures. By monitoring these metrics, timely actions can be taken when the trend indicates a potential breach of the upper control limit.


\section{Conclusion}

According to those exposed throughout the document, it seems clear that from an ethical, normative or regulatory point of view, the monitoring of algorithms based on AI / Machine Learning is an important aspect within their life cycle This allows the detection and correction of  undesirable behaviours of the algorithm, as a consequence of a change in the environmental conditions between the time it was trained and the time of detection of performance or other algorithmic neutrality metrics degradation.

It has been seen that this degradation could come as a consequence of a change in the independent variables or as a consequence of a change in the relationship between the independent variables and the dependent variable. This work collects different ways to measure and monitor metrics to detect trends and notice these degradation gradually over time, allowing to interact with the algorithm consciously during its life cycle.

Within the Maintenance phase of any SW life cycle, it is necessary to execute an incident correction plan, which together with these monitoring techniques, will be complemented with a proactive activity to monitor these Data and Concept Drift metrics that will allow us to anticipate and take preventive measures in order to maintain the solutions within the desired ethical aspects and the current normative or regulatory aspects.


\printbibliography 


\end{document}